\documentclass[11pt, journal, onecolumn]{IEEEtran}
\usepackage[letterpaper, left=1in, right=1in, bottom=1in, top=1in]{geometry}

\usepackage{xcolor,soul,framed} %,caption

\colorlet{shadecolor}{yellow}
\usepackage[pdftex]{graphicx}
\graphicspath{{../pdf/}{../jpeg/}}
\DeclareGraphicsExtensions{.pdf,.jpeg,.png}

\usepackage[cmex10]{amsmath}
\usepackage{array}
\usepackage{mdwmath}
\usepackage{mdwtab}
\usepackage{eqparbox}
\usepackage{url}
\usepackage{listings}
\usepackage{xcolor}
\usepackage{booktabs}
\usepackage{tabularx}
\usepackage{array}
\usepackage[flushleft]{threeparttable}
\usepackage{threeparttablex}
\usepackage{hyperref} % remove boxes with [hidelinks] preamble
\usepackage{lipsum}
\usepackage[ruled,vlined,linesnumbered]{algorithm2e}
\usepackage{footnote}
\usepackage{caption}
\usepackage{subcaption}
\usepackage{amsmath}
\usepackage{amssymb}% for \mathbb
\usepackage{stmaryrd}

\lstdefinestyle{pythonstyle}{
    commentstyle=\color{codegreen},
    keywordstyle=\color{magenta},
    numberstyle=\tiny\color{codegray},
    stringstyle=\color{codepurple},
    basicstyle=\ttfamily\footnotesize,
    breakatwhitespace=false,         
    breaklines=true,                 
    captionpos=b,                    
    keepspaces=true,                 
    numbers=left,                    
    numbersep=5pt,                  
    showspaces=false,                
    showstringspaces=false,
    showtabs=false,                  
    tabsize=2
}

\begin{document}

\renewcommand\IEEEkeywordsname{Keywords} 

\bstctlcite{IEEEexample:BSTcontrol}
   % \title{\textit{Parea}: multi-view hierarchical ensemble clustering for disease subtype discovery}

% TITLE %%%%%%%%%%%%%%%%%%%%%%%
   
%\title {\textit{Pyrea}: multi-view 
% clustering with flexible \\ ensemble structures} %in Python}
\title{\textit{Parea}: multi-view ensemble clustering for cancer subtype discovery}   
%\title {Disease subtype discovery with hierarchical ensemble clustering using \textit{Pyria}}

%%%%%%%%%%%%%%%%%%%%%%%

\DeclareRobustCommand*{\IEEEauthorrefmark}[1]{%
  \raisebox{0pt}[0pt][0pt]{\textsuperscript{\footnotesize\ensuremath{#1}}}}  

\author{\IEEEauthorblockN{Bastian Pfeifer\IEEEauthorrefmark{1}$^{*}$, Marcus D. Bloice\IEEEauthorrefmark{1} and
Michael G. Schimek\IEEEauthorrefmark{1} }

\IEEEauthorblockA{\IEEEauthorrefmark{1}Institute for Medical Informatics, Statistics and Documentation\\ Medical University Graz, Austria}

\thanks{ 
$^{*}$Corresponding author: Bastian Pfeifer (bastian.pfeifer@medunigraz.at).}}

 % \author{Bastian~Pfeifer, Andrei Voicu-Spineanu, Michael G. Schimek~and~Nikolaos Alachiotis}% <-this % stops a space

%\markboth{IEEE BIBM 2021 (\lowercase https://ieeebibm.org/BIBM2021/)}{}

\markboth{Multi-view ensemble clustering}{}

% ====================================================================
\maketitle

% === ABSTRACT ====================================================================
% =================================================================================
\begin{abstract}
%\boldmath
Multi-view clustering methods are essential for the stratification of patients into sub-groups of similar molecular characteristics. In recent years, a wide range of methods has been developed for this purpose. However, due to the high diversity of cancer-related data, a single method may not perform sufficiently well in all cases.
We present Parea$_{\textit{hc}}$, a multi-view hierarchical ensemble clustering approach for disease subtype discovery. We demonstrate its performance on several machine learning benchmark datasets. We apply and validate our methodology on real-world multi-view cancer patient data. Parea$_{\textit{hc}}$ outperforms the current state-of-the-art on six out of seven analysed cancer types. We have integrated the Parea$_{\textit{hc}}$ method into our developed Python package \textit{Pyrea} (\url{https://github.com/mdbloice/Pyrea}), which enables the effortless and flexible design of ensemble workflows while incorporating a wide range of fusion and clustering algorithms. 
\end{abstract}


% === KEYWORDS ====================================================================
% =================================================================================

% \fi
\IEEEpeerreviewmaketitle

% ====================================================================
% ====================================================================
% ====================================================================

% Introduction
\section{Introduction}
\label{sec:introduction}

Multi-view data contain information relevant for the identification of patterns or clusters that allow us to specify groups of subjects or objects. Our focus is on patients for which we have bio-medical and/or clinical observations describing patient characteristics obtained from various diagnostic procedures or produced by different molecular technologies~\cite{fu2020overview}. The different types of subject characteristics constitute views related to these patients.  Integrative clustering of these views facilitates the detection of patient groups, with the consequence of improved clinical diagnostic and treatment schemes.

Simple integration of single view clustering results is not appropriate for the diversity and complexity of available medical observations. Even state-of-the-art multi-view approaches have their limitations. Ensemble clustering has the potential to overcome some of them \cite{ronan2018openensembles}\cite{ alqurashi2019clustering}. For instance, while spectral clustering might be the optimal method for a specific image-based analysis, agglomerative clustering might be more appropriate for tabular data. This can be the case, where patient data reflect some hierarchical structure in a disease of interest and its subtypes \cite{ciriello2013emerging}. Moreover, in real-world applications the data views originate from highly heterogeneous input sources. Thus, each view needs to be clustered with the best possible and most adequate strategy. Multi-view clustering methods are widely applied within the bio-medical domain. Molecular data from different biological layers are retrieved for the same set of patients. The clusters inferred from these multi-omics observations facilitate the stratification of cancer patients into sub-groups, paving the way towards precision medicine. 

Here, we present Parea, a generic and flexible methodology to build clustering ensembles of arbitrary complexity. To be precise, we introduce Parea$_{\textit{hc}}$, an ensemble method which performs hierarchical clustering and data fusion for disease subtype discovery. The name of our method is derived from the Greek word {\it Parea}, meaning a group of friends who gather to share experiences, values, and ideas.

The manuscript is structured as follows: Section~\ref{sec:Parea_general} formally describes the ensemble structures we have developed. Section~\ref{sec:approach} presents our multi-view hierarchical ensemble clustering approach for disease subtype detection. We discuss related work in Section~\ref{sec:other_methods} and introduce the methods we used for benchmark comparisons. In Section~\ref{sec:results} the results are presented and discussed. A brief introduction to the \textit{Pyrea} Python package is given in Section~\ref{sec:pyrea}. We conclude with Section~\ref{sec:conclusion}.

\section{General ensemble architecture}
\label{sec:Parea_general}

The following concept for multi-view ensemble clustering is proposed. Each view $V \in \mathbb{R}^{n\times p}$ is associated with a specific clustering method $c$, where $n$ is the number of samples and $p$ is the number of predictors, and in total we have $N$ data views. An ensemble, called $\mathcal{E}$, can be modelled using a set of views $\mathcal{V}$ and an associated fusion algorithm $f$. 

\begin{equation}
    \mathcal{V} \mapsfrom \{(V \in \mathbb{R}^{n\times p}, c)\} 
\end{equation}

\begin{equation}
    \mathcal{E}(\mathcal{V}, f) \mapsto \widetilde{V}\in \mathbb{R}^{p\times p}
\end{equation}

\begin{equation}
    \mathcal{V} \mapsfrom \{(\widetilde{V}\in \mathbb{R}^{p\times p}, c)\} 
\end{equation}

From the above equations we can see that a specified ensemble $\mathcal{E}$ creates a view $\widetilde{V} \in \mathbb{R}^{p\times p}$ which again can be used to specify $\mathcal{V}$, including an associated clustering algorithm $c$. With this concept it is possible to \textit{layer-wise} stack views and ensembles into arbitrarily complex ensemble architectures. It should be noted, however, that the resulting view of a specified ensemble $\mathcal{E}$ forms an affinity matrix of dimension $p \times p$, and thus only those clustering methods which are congruent with an affinity or a distance matrix for input are applicable. 

% Approach
\section{Proposed Ensemble Approach}
\label{sec:approach}

The Parea$_{\textit{hc}}$ ensemble approach comprises two different strategies: Parea$_{\textit{hc}}^{1}$ is limited to the application of two selected hierarchical clustering methods. Parea$_{\textit{hc}}^{2}$ allows for the hierarchical clustering methods in the data fusion process to be varied. 

\begin{figure*}[h!]
\begin{center}
\includegraphics[width=17.3cm]{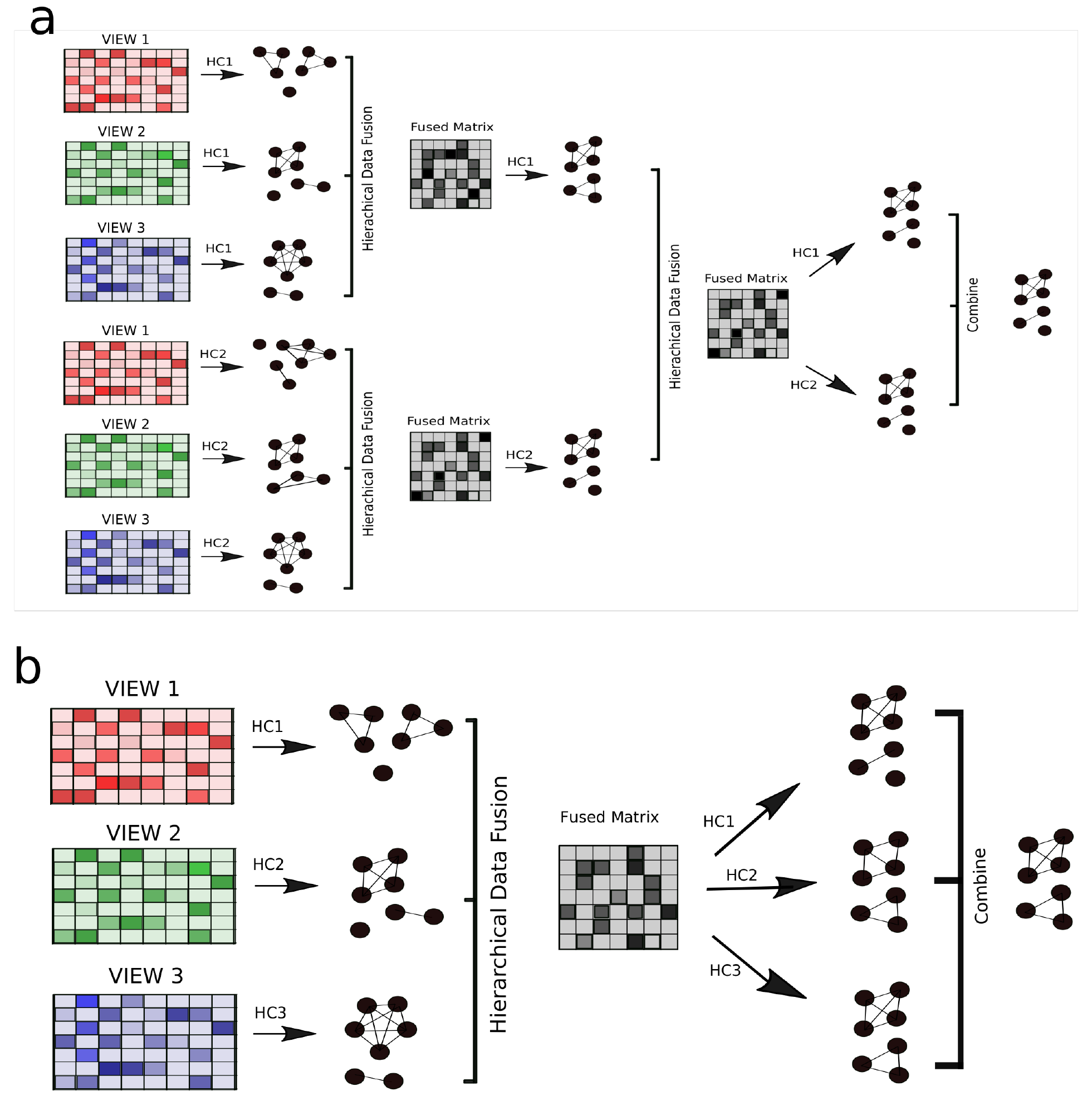}% This is a *.eps file
\end{center}
\caption{\textbf{(a)} The Parea$_{\textit{hc}}^{1}$ ensemble architecture. The views are organised in two ensembles. Each ensemble is associated with a specific clustering method. \textbf{(b)} The Parea$_{\textit{hc}}^{2}$ ensemble architecture. The views can be clustered using different hierarchical clustering techniques. }\label{fig:Parea_arch}
\end{figure*}

The two hierarchical clustering methods of Parea$_{\textit{hc}}^{1}$ for multiple data views are  $hc_{1}$ and $hc_{2}$. The resulting fused matrices $\widetilde{V}$ are clustered again with the same methods and the results are combined to form a final consensus (see Figure \ref{fig:Parea_arch}, panel (a)). A formal description of Parea$_{\textit{hc}}^{1}$ is give by:

\begin{equation}
\mathcal{V}_{1} \mapsfrom \{(V_{1},hc_{1}),(V_{2},hc_{1}),\ldots, (V_{N},hc_{1})\}, 
\quad
\mathcal{V}_{2} \mapsfrom \{(V_{1},hc_{2}),(V_{2},hc_{2}),\ldots, (V_{N},hc_{2})\}
\end{equation}

\begin{equation}
\mathcal{E}_{1}(\mathcal{V}_{1}, f) \mapsto \widetilde{V}_{1},
\quad
\mathcal{E}_{2}(\mathcal{V}_{2}, f) \mapsto \widetilde{V}_{2}
\end{equation}

\begin{equation}
    \mathcal{V}_{3} \mapsfrom \{(\widetilde{V}_{1},hc_{1}),(\widetilde{V}_{2},hc_{2})\} 
\end{equation}

\begin{equation}
\mathcal{E}_{3}(\mathcal{V}_{3}, f) \mapsto \widetilde{V}_{3}.
\end{equation}
The affinity matrix $\widetilde{V}_{3}$ is then clustered with $hc_{1}$ and $hc_{2}$ from the first layer, and the consensus of the obtained clustering solutions constitute the final cluster assignments: 

\begin{equation}
\mathcal{V}_{4} \mapsfrom \{(\widetilde{V_{3}},hc_{1}),(\widetilde{V_{3}},hc_{2})\}
\end{equation}

\begin{equation}
\text{cons}(\mathcal{V}_{4}) 
\end{equation}
Given the proposed ensemble architecture, a genetic algorithm infers the optimal combination of $hc_{1}$ and $hc_{2}$ using the silhouette coefficient \cite{rousseeuw1987silhouettes} as a fitness function. For the data fusion algorithm $f$, we utilize a method introduced in \cite{pfeifer2021hierarchical}. See Figure \ref{fig:Parea_arch} for a graphical illustration of the described ensemble architecture. 

In the Parea$_{\textit{hc}}^{2}$ approach the views are clustered with up to $N$ different hierarchical clustering methods $hc_{1}, hc_{2}, \ldots, hc_{N}$, where $N$ is the number of data views. %{\color{red} SORRY, I CANNOT UNDERSTAND THE FOLLOWING: The fused matrix $\widetilde{V}$ is clustered using the same methods from the first layer and the resulting cluster solutions are finally combined for consensus.} 
A formal description of the Parea$_{\textit{hc}}^{2}$ is given by:

\begin{equation}
\mathcal{V}_{1} \mapsfrom \{(V_{1},hc_{1}),(V_{2},hc_{2}),\ldots, (V_{N},hc_{N})\}
\end{equation}

\begin{equation}
\mathcal{E}_{1}(\mathcal{V}_{1}, f) \mapsto \widetilde{V}_{1}
\end{equation}
The affinity matrix $\widetilde{V}_{1}$ is then clustered with $hc_{1}$, $hc_{2}$, and $hc_{3}$. The consensus of the obtained clustering results are the final cluster assignments:

\begin{equation}
    \mathcal{V}_{2} \mapsfrom \{(\widetilde{V}_{1},hc_{1}),(\widetilde{V}_{1},hc_{2}), (\widetilde{V}_{1},hc_{3})\} 
\end{equation}

\begin{equation}
\text{cons}(\mathcal{V}_{2}) 
\end{equation}
The best combination of clustering methods ($hc_{1}$, $hc_{2}$, and $hc_{3}$) is again inferred by a genetic algorithm, where the silhouette coefficient \cite{rousseeuw1987silhouettes} is deployed as a fitness function. We consider eight different hierarchical clustering methods, namely single-linkage and complete-linkage clustering \cite{murtagh2012algorithms}, an unweighted pair-group method using arithmetic averages (UPGMA) \cite{sokal1958statistical}, a weighted pair-group method using arithmetic averages (WPGMA) \cite{sokal1958statistical} clustering, a weighted pair-group method using centroids (WPGMC) \cite{gower1967comparison}, an unweighted pair-group method using centroids (UPGMC) \cite{sokal1958statistical} clustering, and clustering based on Ward's minimum variance \cite{ward1963hierarchical}\cite{murtagh2014ward}.

% Figure

\section{Alternative approaches}
\label{sec:other_methods}

A comparison with alternative approaches was conducted using a set of state-of-the-art multi-view clustering methods implemented within the Python package mvlearn \cite{perry2021mvlearn}. Part of this set is a multi-view spectral clustering approach with and without the use of a co-training framework \cite{kumar2011co}. An additional method is based on multi-view $k$-means  \cite{chao2017survey} clustering, plus an implementation of multi-view spherical $k$-means using the co-EM framework as described in \cite{bickel2004multi}.

For disease subtype detection we compared Parea$_{\textit{hc}}$ with NEMO \cite{rappoport2019nemo}, SNF (Similar Network Fusion) \cite{wang2014similarity}, HCfused \cite{pfeifer2021hierarchical}, and PINSplus \cite{nguyen2019pinsplus}. SNF models the similarity between subjects or objects as a network and then fuses these networks via an interchanging diffusion process. Spectral clustering is applied to the fused network to infer the final cluster assignments. NEMO builds upon SNF, but provides solutions to partial data and implements a novel \textit{eigen-gap} method \cite{von2007tutorial} to infer the optimal number of clusters.
The method implemented within PINSplus systematically adds noise to the data and infers the best number of clusters based on the stability against this noise. When the best $k$ (number of clusters) is detected, binary  matrices are formulated reflecting the cluster solutions for each single-view contribution. A final agreement matrix is derived by counting the number of times two subjects or objects appear in the same cluster. This agreement matrix is then used for a standard clustering method, such as $k$-means. 

HCfused is a hierarchical clustering and data fusion algorithm. It is based on \textit{bottom-up} agglomerative clustering to create a fused affinity matrix. At each step two clusters are merged within the view that provides the minimal distance between these clusters. The number of times two samples appear in the same cluster is reflected by a co-association matrix. The final cluster assignments are obtained from hierarchical clustering based on Ward's minimum variance \cite{ward1963hierarchical}\cite{murtagh2014ward}.

% Results
\section{Results and discussion}
\label{sec:results}

\subsection*{Evaluation on machine learning benchmark data sets}

\begin{figure}
     \centering
     \begin{subfigure}[b]{0.49\textwidth}
         \centering
         \includegraphics[width=\textwidth]{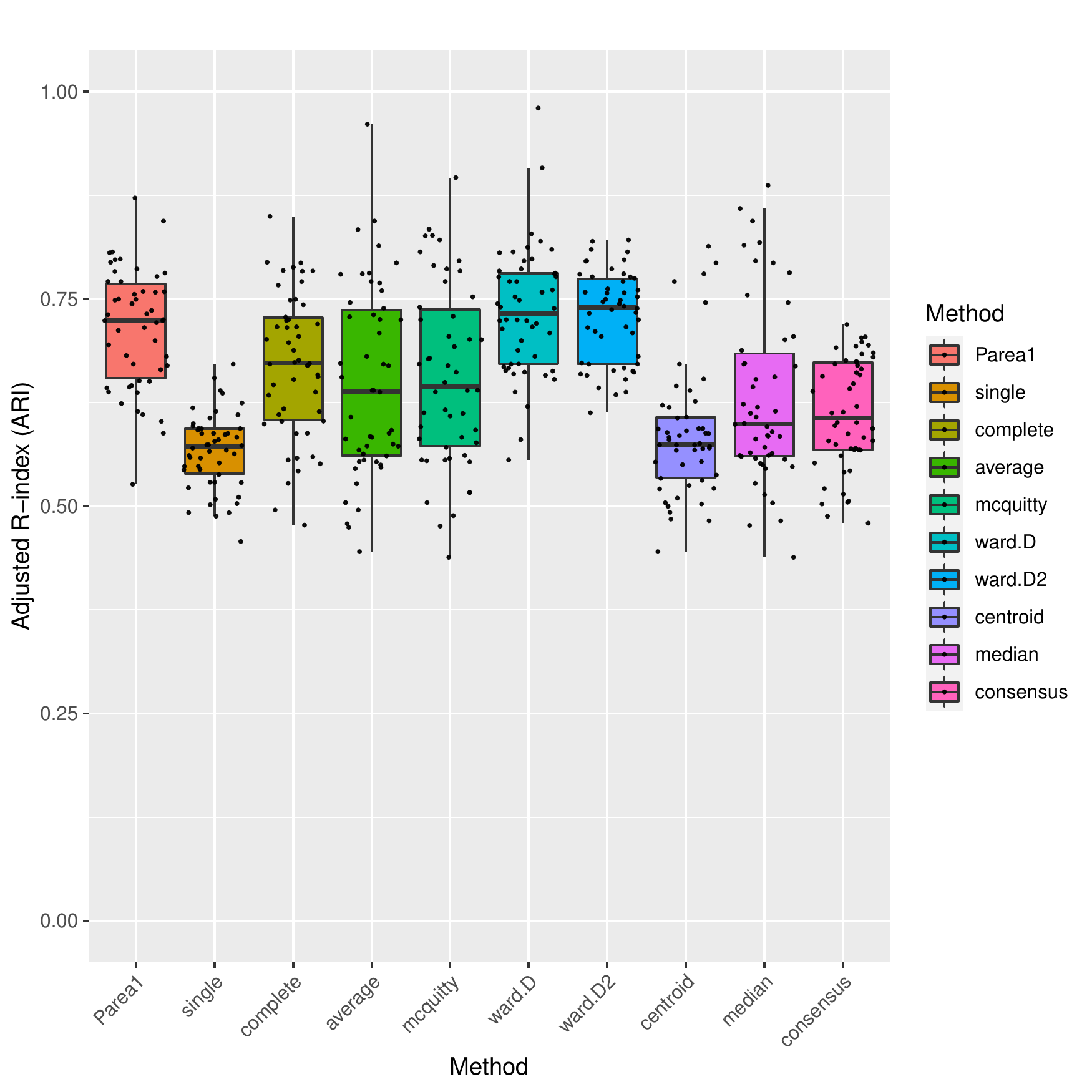}
         \caption{}
        % \label{fig:y equals x}
     \end{subfigure}
     %\hfill
     \begin{subfigure}[b]{0.49\textwidth}
         \centering
         \includegraphics[width=\textwidth]{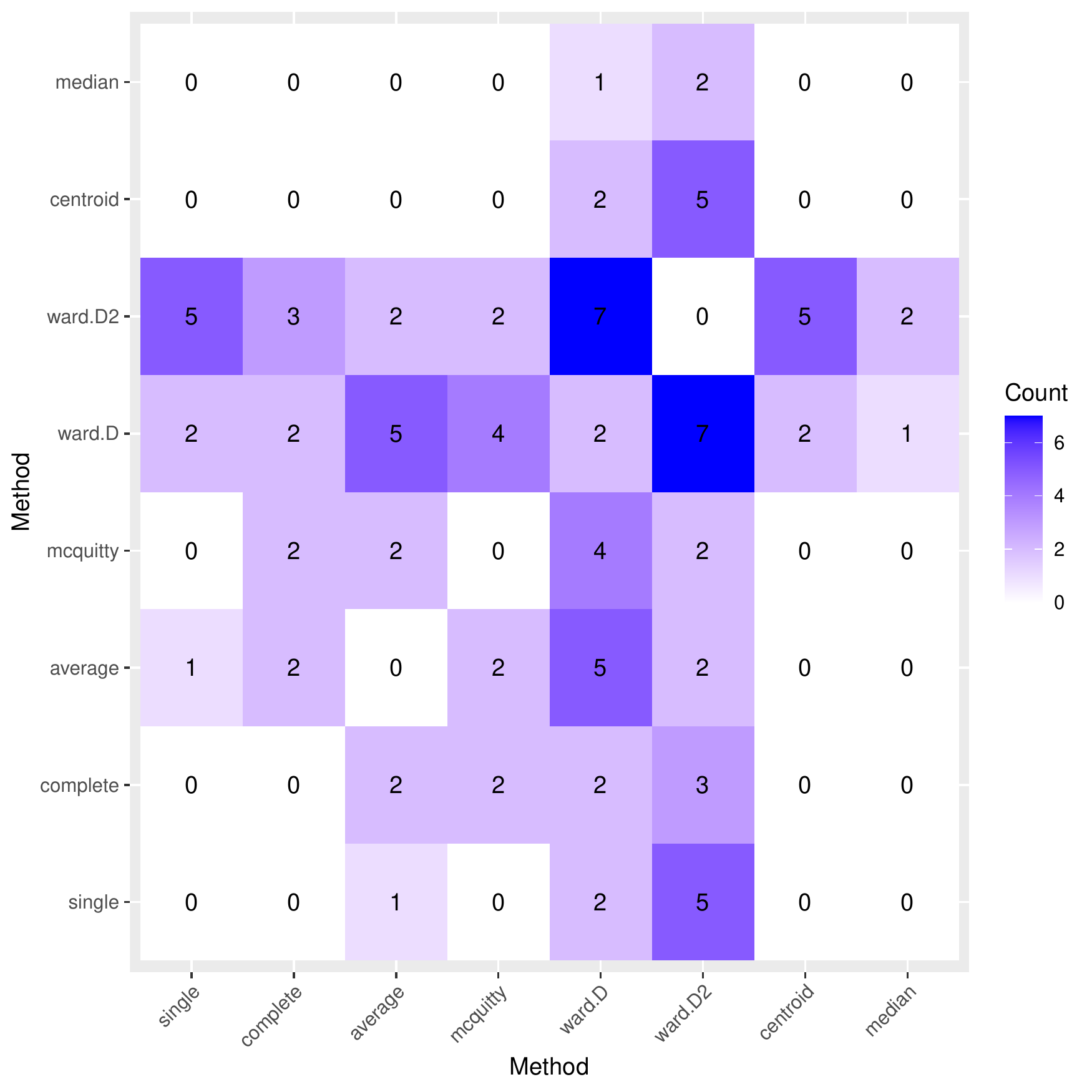}
         \caption{}
         %\label{fig:three sin x}
     \end{subfigure}
        \caption{(a) Parea$_{\textit{hc}}$ versus each of the available ensemble methods executed on the \textit{single-view} IRIS data set. (b) The pairwise Parea$_{\textit{hc}}$ ensembles inferred by the genetic algorithm for clustering the IRIS data set.}
        \label{fig:IRIS}
\end{figure}

We tested Parea$_{\textit{hc}}$ on the IRIS data set available from the UCI Machine Learning Repository\footnote{See \url{https://archive.ics.uci.edu/ml/datasets/iris}}. It contains three classes of 50 instances each, where each class refers to a type of iris plant.
We compared the results of the ensemble with each of the possible ensemble members, and also compared the outcomes with a consensus approach, where the cluster solutions of all single algorithms were combined. The data pool was sampled 50 times and the clustering methods were applied on each iteration. The number of clusters $k$ was set to three corresponding to the ground truth.
These sanity checks revealed superior results for the Parea$_{\textit{hc}}^{1}$ ensemble method compared to a simple consensus approach, as judged by the Adjusted Rand Index (ARI) (see Figure \ref{fig:IRIS}). We could further observe that the underlying genetic algorithm infers ward.D and ward.D2 as the best performing method combination (Figure \ref{fig:IRIS}, panel (b)). 

In an additional investigation, we evaluated the accuracy of the discussed methods on the nutrimouse data set \cite{martin2007novel}. The data set originates from a nutrigenomic study of the mouse in which the effects of five regimens with contrasted fatty acid compositions on liver lipids and hepatic gene expression in mice were considered. Two views were acquired from forty mice. First, gene expressions of 120 genes were measured in liver cells, as potentially relevant for the nutrition study. Second, lipid concentrations of 21 hepatic fatty acids were measured by gas chromatography.

\begin{figure}
     \centering
     \begin{subfigure}[b]{0.49\textwidth}
         \centering
         \includegraphics[width=\textwidth]{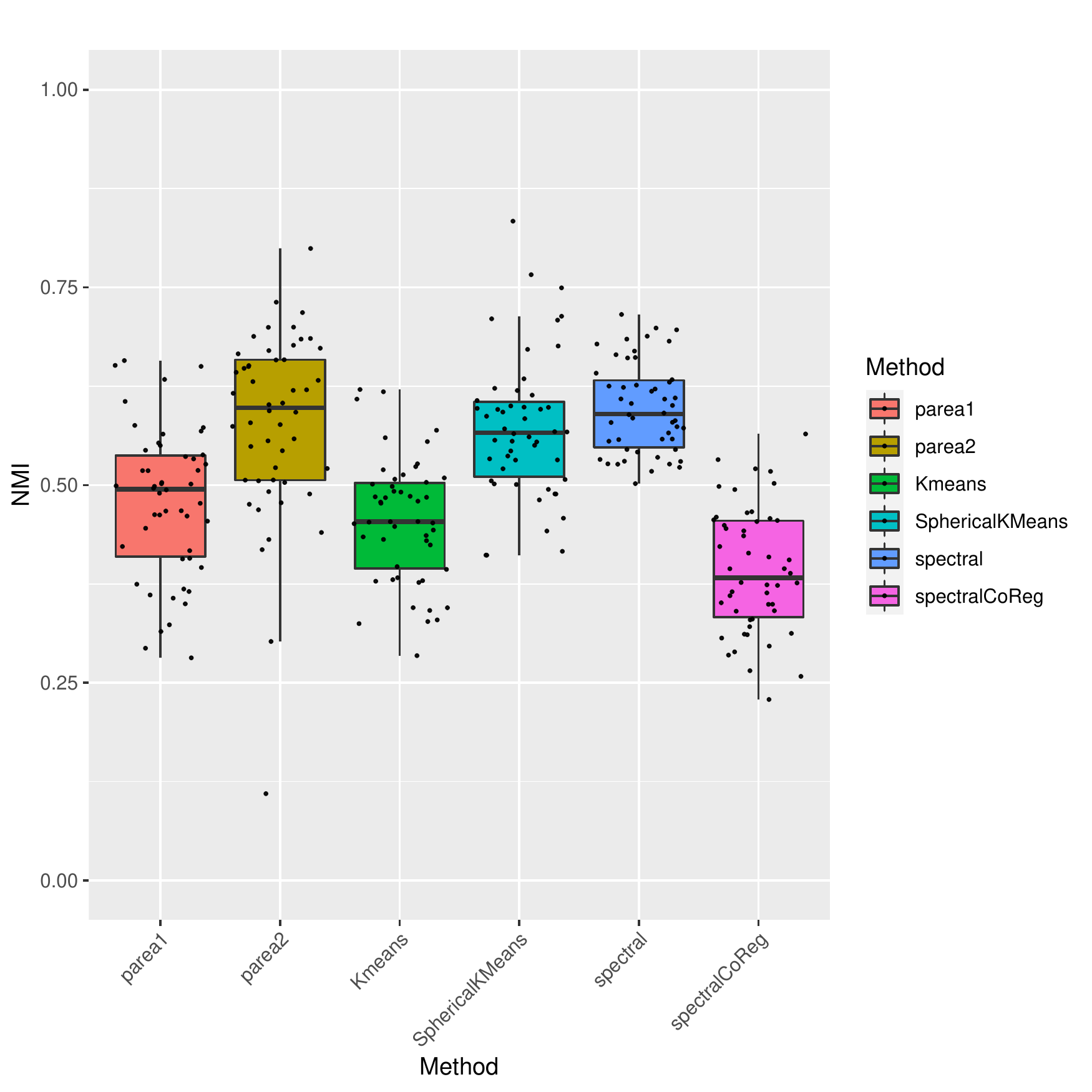}
         \caption{}
        % \label{fig:y equals x}
     \end{subfigure}
     %\hfill
     \begin{subfigure}[b]{0.50\textwidth}
         \centering
         \includegraphics[width=\textwidth]{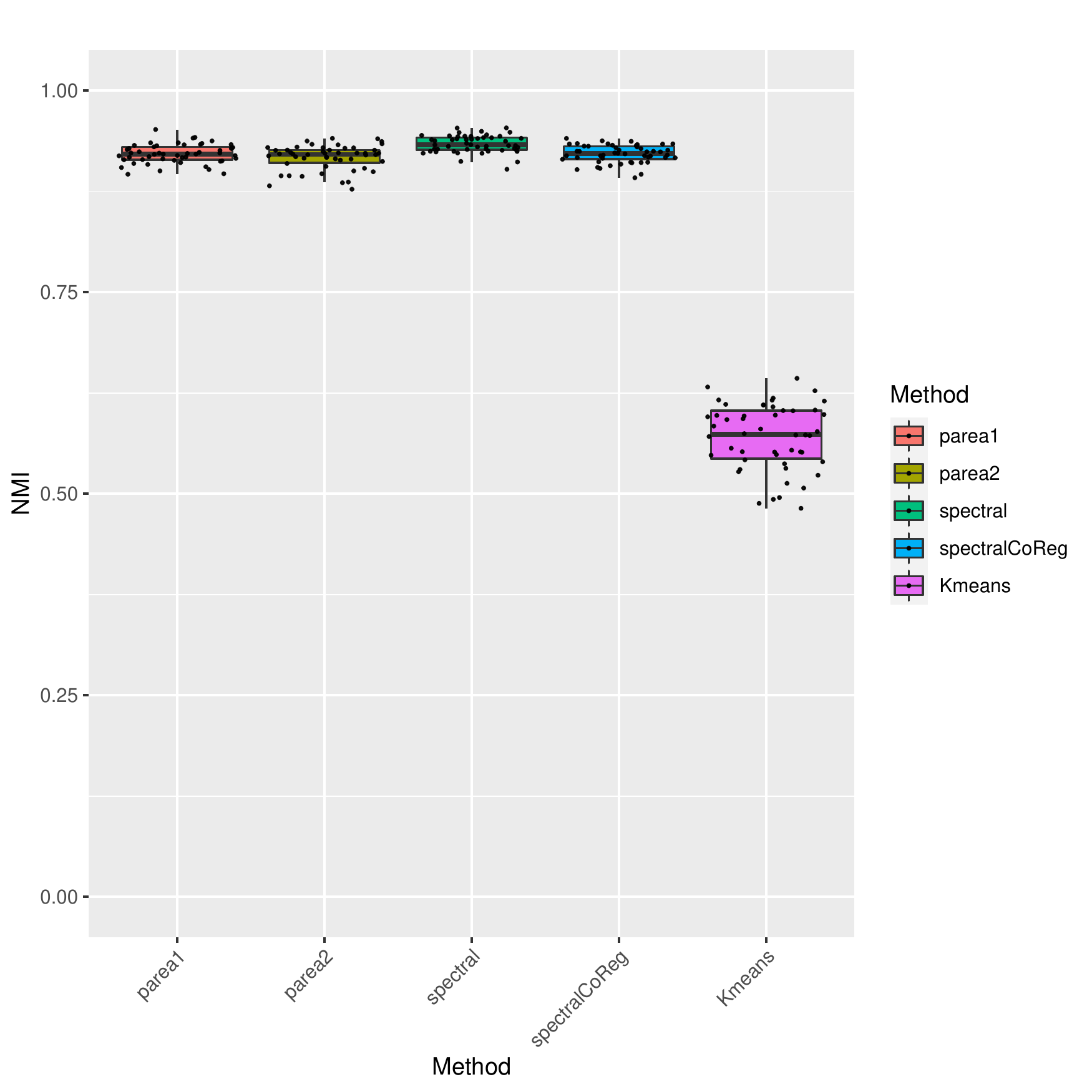}
         \caption{}
         %\label{fig:three sin x}
     \end{subfigure}
        \caption{
        Parea$_{\textit{hc}}$ versus a set of state-of-the-art multi-view methods implemented within the mvlearn package \cite{perry2021mvlearn}.
        (a) The \textit{multi-view} nutrimouse data set. (b) The \textit{multi-view} 100 leaves data set.
       }
        \label{fig:100leaves}
\end{figure}

For the nutrimouse data set Parea$_{\textit{hc}}^{2}$ performs better than Parea$_{\textit{hc}}^{1}$ (see Figure \ref{fig:100leaves}, panel (a)). This observation suggests that higher accuracy can be achieved when the views are analysed with disjoint clustering strategies. Parea$_{\textit{hc}}^{2}$ performs best when the median NMI (\textit{Normalised Mutual Information}) is used as a metric. However, at the same time we observed higher variance for the Parea ensembles compared to multi-view spectral clustering.  It is worth noting that the alternative spectral-based approaches from the mvlearn Python package performed as well as Parea$_{\textit{hc}}^{1}$. 

Last, we further studied Parea's performance on the one-hundred plant species leaves multi-view data set, available from the UCI Machine Learning Repository\footnote{See \url{https://archive.ics.uci.edu/ml/datasets/One-hundred+plant+species+leaves+data+set}}. For each feature, a 64 element vector is observed per leaf sample. These vectors form contiguous descriptors for shape as well as texture and margin. In contrast to the IRIS data set it is composed of multiple views. As can be seen in Figure \ref{fig:100leaves}, Parea$_{\textit{hc}}^{1}$ and Parea$_{\textit{hc}}^{2}$ compete well with the spectral-based approaches. Multi-view $k$-means clustering does not capture the ground truth class distribution (see Figure \ref{fig:100leaves}, panel (b)).

%%%% IRIS
% END OF IRIS

%%% ML datasets
% END OF ML datasets

\subsection*{Multi-omics clustering for disease subtype discovery}

We applied the aforementioned ensemble methods to patient data of seven different cancer types, namely glioblastoma multiforme (GBM), kidney renal clear cell carcinoma (KIRC), liver hepatocellular carcinoma (LIHC), skin cutaneous melanoma (SKCM), ovarian serous cystadenocarcinoma (OV), sarcoma (SARC), and acute myeloid leukemia (AML), aiming at the externally known survival outcome. The Parea$_{\textit{hc}}$ ensemble approach was studied on multi-omics data, including gene expression data (mRNA), DNA methylation, and micro-RNAs. The obtained results were compared with the alternative approaches SNF, NEMO, PINSplus, and HCfused. All data were retrieved from the ACGT lab at Tel Aviv University\footnote{See \url{http://acgt.cs.tau.ac.il/multi_omic_benchmark/download.html}}, which makes available a data repository recently proposed as a convenient benchmark for multi-omics clustering approaches \cite{rappoport2018multi}. The data were pre-processed as follows: patients and features with more than $20\%$ missing values were removed and the remaining missing values were imputed with $k$-nearest neighbor imputation. In the methylation data, we selected those 5000 features with maximal variance in each data set. All features were then normalised to have mean zero and standard deviation one. 

The resulting multi-omics views were then clustered with the discussed multi-view clustering approaches. It is important to mention that the cancer patients were exclusively analysed based on their genomic footprints. The survival statuses and times of all patients were retrieved to validate the quality of the inferred patient clusters. Here, a well performing clustering method is determined by its capability to separate patients into groups with similar event statuses in terms of survival. To this end we randomly sampled 100 patients 30 times from the data pool and performed the Cox log-rank test, which is an inferential procedure for the comparison of event time distributions among independent (i.e. clustered) patient groups.

\begin{figure}[h!]
\begin{center}
\includegraphics[width=14cm]{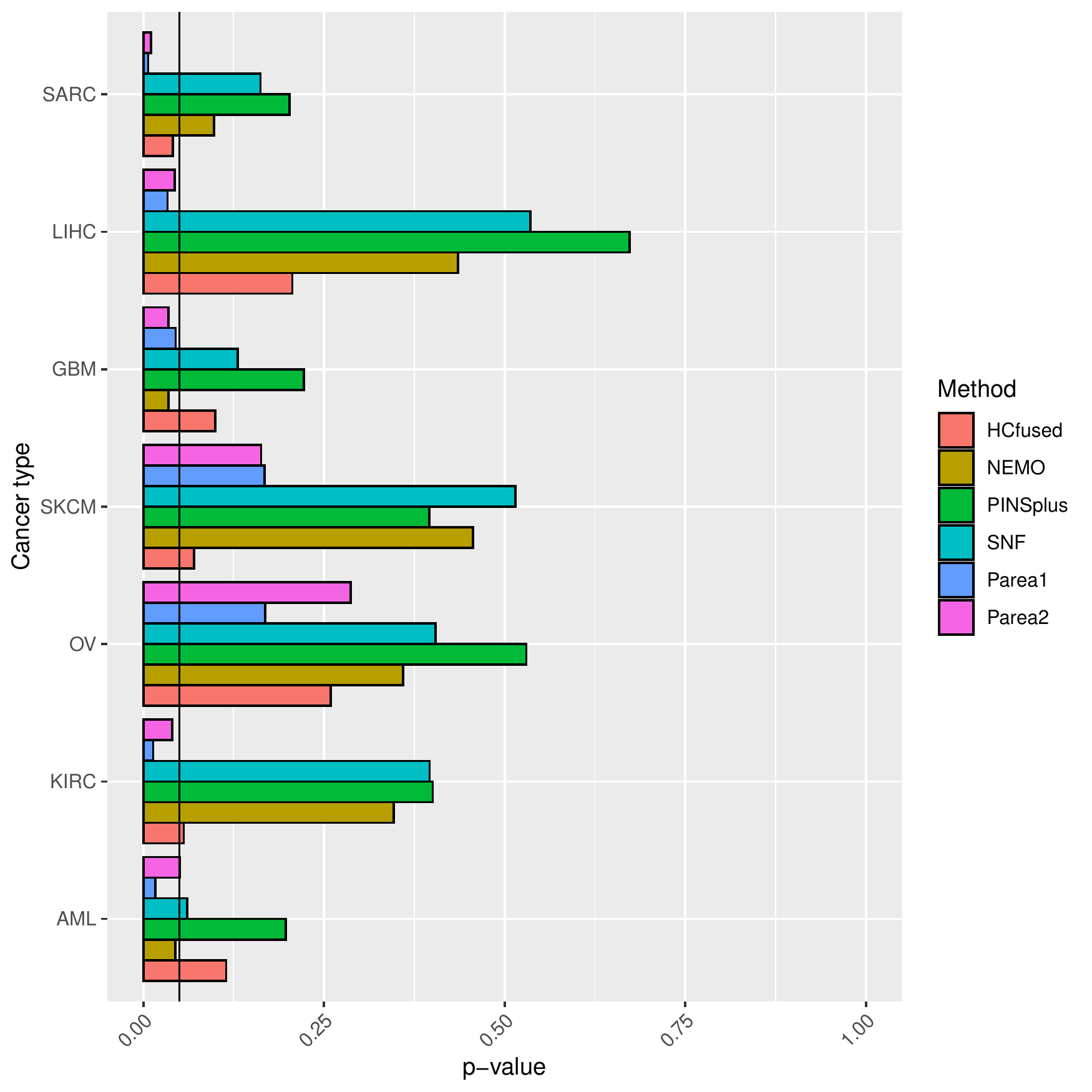}% This is a *.eps file
\end{center}
\caption{Parea$_{\textit{hc}}$ in comparison with the alternative approaches SNF, NEMO, PINSplus, and HCfused. Colored bars represent the method-specific median Cox log-rank $p$-values for the seven different cancer types. The vertical line refers to $\alpha=0.05$ significance level.}
\label{fig:TCGA}
\end{figure}

In the case of Parea$_{\textit{hc}}$, the optimal number of clusters was determined by the silhouette coefficient. We set the number of HCfused \cite{pfeifer2021hierarchical} fusion iterations to 30. The maximum possible number of clusters was fixed to 10. The obtained Cox log-rank $p$-values ($\alpha=0.05$ significance level) are displayed in Figure \ref{fig:TCGA} and Table \ref{tab:table1}. Our Parea$_{\textit{hc}}$ ensembles outperformed the alternative approaches in six out of seven cases. SKCM is the only cancer type for which HCfused achieved better results. Notably, spectral-based clustering (see NEMO \cite{rappoport2019nemo} and SNF \cite{wang2014similarity}) does not perform as well as with the benchmark machine learning data sets. We further learned that Parea$_{\textit{hc}}^{1}$ is more accurate than Parea$_{\textit{hc}}^{2}$ (Figure \ref{fig:TCGA}, Table \ref{tab:table1}).  

% TCGA %%%%%%%%%%%%%%%%%%%%%%%%%%%%%%%%%%%%%%%
% END OF TCGA
%%%%%%%%%%%%%%%%%%%%%%%%%%%%%%%%%%%%%%%%%%%%%

%\begin{figure}
%     \centering
%     \begin{subfigure}[b]{0.49\textwidth}
%         \centering
%         \includegraphics[width=\textwidth]{Figures/Pare%         \caption{}
%        % \label{fig:y equals x}
%     \end{subfigure}
%     %\hfill
%     \begin{subfigure}[b]{0.50\textwidth}
%         \centering
%         \includegraphics[width=\textwidth]{Figures/Parea_SE.pdf}
%         \caption{}
         %\label{fig:three sin x}
%     \end{subfigure}
%        \caption{(a) Parea$_{\textit{hc}}$ in comparison with the alternative methods SNF, NEMO, PINSplus, and HCfused. Boxplots of the Cox log-rank $p$-values (on a negative log10 scale) are displayed for the seven different cancer types. The vertical line refers to $\alpha=0.05$ significance level. The higher the log-scaled $p$-values the better. (b) The mean Cox log-rank $p$-values and the corresponding standard deviation as error bars.}
       % \label{fig:TCGA}
%\end{figure}

\begin{table}[h!]
  \begin{center}
    \caption{Survival analysis of TCGA cancer data}
    \label{tab:table1}
    \begin{tabular}{l l l l l l l l} % <-- Alignments: 1st column left, 2nd middle and 3rd right, with vertical lines in between
      \textbf{Cancer type} & \textbf{Sample size} & SNF & PINSplus & NEMO & HCfused &Parea$_{\textit{hc}}^{1}$ & Parea$_{\textit{hc}}^{2}$ \\
      \hline
      GBM & 538 & 0.1304 & 0.2223 & \textbf{0.0347} & 0.0997 & \textbf{0.0447}& \textbf{0.0347}\\
      KIRC & 606 & 0.3962 & 0.4005 & 0.3464 & 0.0561 & \textbf{0.0137} & \textbf{0.0400}\\
     % COAD & 328 & 0.6402 & \textbf{0.4500} & 0.6092 & 0.7081 &\\
      LIHC & 423 & 0.5357 & 0.6731 & 0.4354 & 0.2062 & \textbf{0.0334} & \textbf{0.0436}\\
      SKCM & 473 & 0.5153 & 0.3956 & 0.4565 & 0.0699 & 0.1677 & 0.1629\\
      OV & 307 & 0.4042 & 0.5300 & 0.3593 & 0.2594 & 0.1685 & 0.2870 \\
      SARC & 265 & 0.1622 & 0.2024 & 0.0979 & \textbf{0.0408} & \textbf{0.0076} & \textbf{0.0109}  \\
      AML & 173 & 0.0604 & 0.1973 & \textbf{0.0440} & 0.1148 & \textbf{0.0167} & 0.0502\\ \hline
      %BIC & 1212 & 0.3004 & 0.6836 & \textbf{0.1771} & 0.1870 &\\ \hline
    \end{tabular}
  \end{center}
  \centering
Display of the median Cox log-rank p-values. Significant results ($\alpha=0.05$) are highlighted in bold. 
\end{table}

\section{Pyrea Software Library}
\label{sec:pyrea}
Pyrea is a software framework which allows for flexible ensembles to be built, layer-wise, using a variety of clustering algorithms and fusion techniques. It can be used to implement the ensemble methods discussed throughout this paper and any other custom ensemble structure that users may wish to create. It is made available as a Python package, which can be installed via the \textit{pip} package manager. See the project's GitHub repository under \url{https://github.com/mdbloice/Pyrea} for source code and usage examples, such as how to implement some of the ensembles mentioned above. Comprehensive documentation is available under \url{https://pyrea.readthedocs.io}. Pyrea is MIT licensed, and available for Windows, macOS, and Linux.

\section{Conclusion}
\label{sec:conclusion}
We have presented Parea$_{\textit{hc}}$, an ensemble approach for multi-view hierarchical clustering for cancer subtype discovery. We could show that Parea$_{\textit{hc}}$ competes well with current state-of-the-art multi-view clustering techniques, on classical machine learning data sets as well as for real-world multi-omics cancer patient data. The proposed methodology for building ensembles is highly versatile and allows for ensembles to be stacked layer-wise. Additionally, the Parea ensemble strategy is not limited to a specific clustering technique. Within our developed Python package \textit{Pyrea}, we enable flexible ensemble building, while providing a wide-range of clustering algorithms, data fusion techniques, and metrics to infer the best number of clusters.

\bibliographystyle{IEEEtran}
\bibliography{IEEEabrv,Bibliography}

\begin{thebibliography}{10}
\providecommand{\url}[1]{#1}
\csname url@rmstyle\endcsname
\providecommand{\newblock}{\relax}
\providecommand{\bibinfo}[2]{#2}
\providecommand\BIBentrySTDinterwordspacing{\spaceskip=0pt\relax}
\providecommand\BIBentryALTinterwordstretchfactor{4}
\providecommand\BIBentryALTinterwordspacing{\spaceskip=\fontdimen2\font plus
\BIBentryALTinterwordstretchfactor\fontdimen3\font minus
  \fontdimen4\font\relax}
\providecommand\BIBforeignlanguage[2]{{%
\expandafter\ifx\csname l@#1\endcsname\relax
\typeout{** WARNING: IEEEtran.bst: No hyphenation pattern has been}%
\typeout{** loaded for the language `#1'. Using the pattern for}%
\typeout{** the default language instead.}%
\else
\language=\csname l@#1\endcsname
\fi
#2}}

\bibitem{fu2020overview}
L.~Fu, P.~Lin, A.~V. Vasilakos, and S.~Wang, ``An overview of recent multi-view
  clustering,'' \emph{Neurocomputing}, vol. 402, pp. 148--161, 2020.

\bibitem{ronan2018openensembles}
T.~Ronan, S.~Anastasio, Z.~Qi, R.~Sloutsky, K.~M. Naegle, and P.~H. S.~V.
  Tavares, ``Openensembles: a python resource for ensemble clustering,''
  \emph{The Journal of Machine Learning Research}, vol.~19, no.~1, pp.
  956--961, 2018.

\bibitem{alqurashi2019clustering}
T.~Alqurashi and W.~Wang, ``Clustering ensemble method,'' \emph{International
  Journal of Machine Learning and Cybernetics}, vol.~10, no.~6, pp. 1227--1246,
  2019.

\bibitem{ciriello2013emerging}
G.~Ciriello, M.~L. Miller, B.~A. Aksoy, Y.~Senbabaoglu, N.~Schultz, and
  C.~Sander, ``Emerging landscape of oncogenic signatures across human
  cancers,'' \emph{Nature Genetics}, vol.~45, no.~10, pp. 1127--1133, 2013.

\bibitem{rousseeuw1987silhouettes}
P.~J. Rousseeuw, ``Silhouettes: a graphical aid to the interpretation and
  validation of cluster analysis,'' \emph{Journal of Computational and Applied
  Mathematics}, vol.~20, pp. 53--65, 1987.

\bibitem{pfeifer2021hierarchical}
B.~Pfeifer and M.~G. Schimek, ``A hierarchical clustering and data fusion
  approach for disease subtype discovery,'' \emph{Journal of Biomedical
  Informatics}, vol. 113, p. 103636, 2021.

\bibitem{murtagh2012algorithms}
F.~Murtagh and P.~Contreras, ``Algorithms for hierarchical clustering: an
  overview,'' \emph{Wiley Interdisciplinary Reviews: Data Mining and Knowledge
  Discovery}, vol.~2, no.~1, pp. 86--97, 2012.

\bibitem{sokal1958statistical}
R.~R. Sokal, ``A statistical method for evaluating systematic relationships.''
  \emph{Univ. Kansas, Sci. Bull.}, vol.~38, pp. 1409--1438, 1958.

\bibitem{gower1967comparison}
J.~C. Gower, ``A comparison of some methods of cluster analysis,''
  \emph{Biometrics}, pp. 623--637, 1967.

\bibitem{ward1963hierarchical}
J.~H. Ward~Jr, ``Hierarchical grouping to optimize an objective function,''
  \emph{Journal of the American Statistical Association}, vol.~58, no. 301, pp.
  236--244, 1963.

\bibitem{murtagh2014ward}
F.~Murtagh and P.~Legendre, ``Ward’s hierarchical agglomerative clustering
  method: which algorithms implement ward’s criterion?'' \emph{Journal of
  Classification}, vol.~31, no.~3, pp. 274--295, 2014.

\bibitem{perry2021mvlearn}
R.~Perry, G.~Mischler, R.~Guo, T.~Lee, A.~Chang, A.~Koul, C.~Franz, H.~Richard,
  I.~Carmichael, P.~Ablin, \emph{et~al.}, ``mvlearn: Multiview machine learning
  in python.'' \emph{J. Mach. Learn. Res.}, vol.~22, pp. 109--1, 2021.

\bibitem{kumar2011co}
A.~Kumar and H.~Daum{\'e}, ``A co-training approach for multi-view spectral
  clustering,'' in \emph{Proceedings of the 28th international conference on
  machine learning (ICML-11)}.\hskip 1em plus 0.5em minus 0.4em\relax Citeseer,
  2011, pp. 393--400.

\bibitem{chao2017survey}
G.~Chao, S.~Sun, and J.~Bi, ``A survey on multi-view clustering,'' \emph{arXiv
  preprint arXiv:1712.06246}, 2017.

\bibitem{bickel2004multi}
S.~Bickel and T.~Scheffer, ``Multi-view clustering.'' in \emph{ICDM}, vol.~4,
  no. 2004.\hskip 1em plus 0.5em minus 0.4em\relax Citeseer, 2004, pp. 19--26.

\bibitem{rappoport2019nemo}
N.~Rappoport and R.~Shamir, ``{NEMO}: cancer subtyping by integration of
  partial multi-omic data,'' \emph{Bioinformatics}, vol.~35, no.~18, pp.
  3348--3356, 2019.

\bibitem{wang2014similarity}
B.~Wang, A.~M. Mezlini, F.~Demir, M.~Fiume, Z.~Tu, M.~Brudno, B.~Haibe-Kains,
  and A.~Goldenberg, ``Similarity network fusion for aggregating data types on
  a genomic scale,'' \emph{Nature Methods}, vol.~11, no.~3, pp. 333--337, 2014.

\bibitem{nguyen2019pinsplus}
H.~Nguyen, S.~Shrestha, S.~Draghici, and T.~Nguyen, ``{PINSplus}: a tool for
  tumor subtype discovery in integrated genomic data,'' \emph{Bioinformatics},
  vol.~35, no.~16, pp. 2843--2846, 2019.

\bibitem{von2007tutorial}
U.~von Luxburg, ``A tutorial on spectral clustering,'' \emph{Statistics and
  Computing}, vol.~17, no.~4, pp. 395--416, 2007.

\bibitem{martin2007novel}
P.~G. Martin, H.~Guillou, F.~Lasserre, S.~D{\'e}jean, A.~Lan, J.-M. Pascussi,
  M.~SanCristobal, P.~Legrand, P.~Besse, and T.~Pineau, ``Novel aspects of
  ppar$\alpha$-mediated regulation of lipid and xenobiotic metabolism revealed
  through a nutrigenomic study,'' \emph{Hepatology}, vol.~45, no.~3, pp.
  767--777, 2007.

\bibitem{rappoport2018multi}
N.~Rappoport and R.~Shamir, ``Multi-omic and multi-view clustering algorithms:
  review and cancer benchmark,'' \emph{Nucleic Acids Research}, vol.~46,
  no.~20, pp. 10\,546--10\,562, 2018.

\end{thebibliography}

\vfill


\end{document}